\documentclass{article}

\usepackage{PRIMEarxiv}

\usepackage{amsmath,amsthm,amssymb,amsfonts}
\usepackage{bm}
\usepackage{algorithm, algorithmic}%
\usepackage{graphicx}
\usepackage{textcomp}
\usepackage{xcolor}

\usepackage{enumerate} % Pour mieux gérer la commande enumerate dans les sections
\usepackage{units}	% Pour pouvoir tapper les unités correctement
\usepackage{subcaption}

\usepackage{diagbox}
\usepackage{multirow}

\title{Multi-Objective Algorithms for Learning Open-Ended Robotic Problems 
}

\author{
  Martin Robert \\
  Dept. of Math and Informatics \\
  Universite De Sherbrooke \\
  Sherbrooke, Canada \\
  \texttt{martin.robert2@USherbrooke.ca} \\
   \And
  Simon Brodeur \\
  Menya AI division \\
  Levio \\
  Sherbrooke, Canada\\
  \texttt{simon.brodeur@menya.levio.ca} \\
  \AND
  Francois Ferland \\
  Dept. of Electrical Eng. and Computer Eng. \\
  Universite De Sherbrooke \\
  Sherbrooke, Canada \\
  \texttt{Francois.Ferland@USherbrooke.ca} \\
}

\begin{document}

\maketitle

\begin{abstract}
Quadrupedal locomotion is a complex, open-ended problem vital to expanding autonomous vehicle reach. Traditional reinforcement learning approaches often fall short due to training instability and sample inefficiency. We propose a novel method leveraging multi-objective evolutionary algorithms as an automatic curriculum learning mechanism, which we named Multi-Objective Learning (MOL). Our approach significantly enhances the learning process by projecting velocity commands into an objective space and optimizing for both performance and diversity. Tested within the MuJoCo physics simulator, our method demonstrates superior stability and adaptability compared to baseline approaches. As such, it achieved 19\% and 44\% fewer errors against our best baseline algorithm in difficult scenarios based on a uniform and tailored evaluation respectively. This work introduces a robust framework for training quadrupedal robots, promising significant advancements in robotic locomotion and open-ended robotic problems.
\end{abstract}

\section{Introduction}

%%%%% Classic - RL approach to quadruped locomotion %%%%%
Improving autonomous vehicle mobility and reach with quadrupedal locomotion traditionally relied on classic approaches, such as optimal control and planning, necessitated intricate modelling of robotic dynamics and demanded significant computational time \cite{chignoli2021humanoid}. Avoiding these shortcomings, Reinforcement Learning (RL) has become a state-of-the-art approach \cite{miki2022learning}. However, it is still vulnerable to generating unpredictable behaviours, has issues with reproducibility, training instability, difficulties in balancing exploration and exploitation, and sample inefficiency \cite{andrychowicz2021matters}.

%%%%% Open-Ended Problems %%%%%
Moreover, as animals do when striving for natural locomotion in the real world, we encounter an open-ended problem due to the limitless ways to achieve locomotion. In quadruped locomotion, no single walking gait allows movement in all directions. Thus, a locomotion controller must learn diverse behaviours for various locomotion tasks, with no specific behaviour being universally superior. Hence, open-endness makes quadrupedal robotic locomotion particularly challenging, even for RL.

%%%%% Automatic Curriculum Learning %%%%%
Humans also encounter such challenges. Consequently, curriculum learning has emerged as a pivotal mechanism in human education, aiding the acquisition of novel tasks. However, the manual creation of curricula can be time-consuming. Automatic Curriculum Learning (ACL) approaches partially mitigate this problem. While current ACL approaches to open-ended robotic tasks demonstrated novelty and improvement \cite{ijcai2020p671}, we argue that they may be computationally demanding, complex, brittle, and lack a conjoint directed effort toward higher performance and diversity.

\begin{figure}
  \centering
  \includegraphics[width=0.3\linewidth]{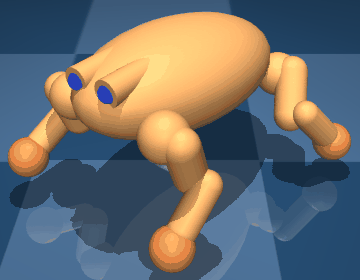}
\caption{Render of the simulated 12 degrees of freedom quadrupedal robot used in the MuJuCo environment.}
\label{fig:robot_dog}
\end{figure}

%%%%% specify task space and proposed solution %%%%%
This work addresses the challenge of developing a quadrupedal locomotion controller that achieves desired walking velocities. We conditioned the controller on the x and y linear velocities and z angular velocity, with each combination of velocities regarded as a task, which we refer to as a velocity command. Given the problem's complexity, the achievable range of commands, and the numerous optimal walking gaits, we consider this an open-ended problem. 

We propose to look at quadrupedal locomotion as a Multi-Objective (MO) problem where trade-offs are inevitable. We thus suggest projecting commands into an objective space and using the achieved performance for each command as a magnitude within this space. A MO algorithm can then select commands that enhance performance and diversity, thereby improving the controller's learning. Our approach, that we named Multi-Objective Learning (MOL), is tested on a quadrupedal robot with 12 degrees of freedom simulated in the MuJoCo physics simulator \cite{todorov2012mujoco}. A render of the robot is shown in Figure \ref{fig:robot_dog}. While this design was not modeled after an actual robot, we believe a realistically design robot in MuJoCo could learn with our approach and transfer to a real-world robot, such as in \cite{haarnoja2023learning}. Additionally, in \cite{miki2022learning}, they showed the effectiveness of a teacher-student approach in simulation to enable the use of the learned controller on a real-world robot.

%%%%% contribution %%%%%
This work presents three key contributions in the domain of ACL for quadrupedal locomotion:

\begin{itemize}
\item Introduction of the novel MOL approach to adapt MO evolutionary algorithms for ACL in a continuous task space.
\item Generation of a diverse set of high-performing commands for quadrupedal locomotion, maintaining or enhancing performance across the entire command space.
\item Enhancement of the stability of learning quadrupedal locomotion in complex scenarios within the realistic MuJoCo physics simulator. 
\end{itemize}
 
\section{Related Work}\label{related_work}
Beginning with the seminal work of \cite{bengio2009curriculum}, Curriculum Learning is depicted as a mechanism that systematically increases task complexity, thereby optimizing training convergence and effectively navigating non-convex optimization landscapes. A step further is automatically generating the curriculum by discovering how to increase task complexity.

A parallel to the Zone of Proximal Development (ZPD) \cite{shabani2010vygotsky} from psychology is explored in \cite{seita2019zpd}, who apply it to deep reinforcement learning, tailoring task difficulty to an agent's evolving capabilities by choosing teacher policies of adequate skill level. Similarly, \cite{tzannetos2023proximal} also makes use of the concept of ZDP, but focuses on proximally adjusting the curriculum’s complexity with an approach called probability of success which is based on the agent's current ability. In \cite{huizinga2022evolving}, the impact of sampling combinations of tasks from a set of discrete tasks is explored to serve as gradual stepping stones to reach a solution solving the whole set of tasks. They relied on an MO algorithm to select the best performing and behaviorally diverse individuals on each set of discrete tasks. This differs from our approach, which uses MO algorithms to select tasks, is applied to more than two objectives and uses all simulations experience to update our controller instead of discarding individuals. A different approach to building curriculums is to use opponents of similar strength, as such \cite{sukhbaatar2018intrinsic} introduce asymmetric self-play between dual agents, to generate and adjust task complexities dynamically.

While these approaches are interesting, they focus on a set number of well-defined tasks, but in this work, we focused on ACL employed for open-ended problems. As such, the POET system investigates the use of co-evolution to evolve both new environment tasks and new agents to solve them \cite{wang2019paired}. POET simultaneously generates new behaviours and tasks by optimizing each agent to a new environment, but this also means that trained agents are discarded, which decreases efficiency. In \cite{mehta2020active}, they introduce Active Domain Randomization, where instead of randomly selecting the randomization parameters, it sets the problem of finding a challenging set of parameters as an RL problem where a discriminator network provides the reward. Similarly, \cite{akkaya2019solving} presents Automatic Domain Randomization (ADR) for training a robotic hand to solve the Rubik's cube. It does so by gradually increasing the sampling range of the physics parameters based on the agent's current success. Additionally, the work by \cite{pmlr-v80-florensa18a} on Automatic Goal Generation uses Generative Adversarial Networks (GAN) to generate goals that are in the set of goals of intermediate difficulty and discriminate the ones that are not. Close to this, \cite{DBLP-conf-iclr-RacaniereLSRFL20} proposes to use setter-solver interactions, which is tightly related to the GAN approach, but they extend the definition of the valid space of goals to consider. Lastly, \cite{wang2023toward} proposed to use a diffusion model to generate plans for new unseen tasks, while being flexible about the actor executing the plan. As described here, current approaches applicable to open-ended problems do not look at them as a space of objectives with varying degrees of difficulty and trade-offs. We thus applied the MOL approach to challenging scenarios and were shown to be more efficient at exploring and understanding the task space. We attribute this to evolutionary MO algorithms, which enable reaching a more diverse range of tasks in which a robot performs well.

%%%%%%%%%% Work Relevance %%%%%%%%%%
As discussed, approaches to open-ended robotic tasks include, but are not limited to, the co-evolution of agents and environments \cite{wang2019paired}, which must handle high computational demand. Then, there is the progressive increase of task range based on current progress \cite{akkaya2019solving}, which cannot explore non-linear task space. Lastly, the use of a goal generator network and a discriminator network \cite{pmlr-v80-florensa18a, DBLP-conf-iclr-RacaniereLSRFL20}, which come with the complexity and variance of the GAN convergence ability. While using mutation and crossover parameters to adjust task difficulty, MOL employs a gradient-free local search method to reduce training variance. It mitigates high computational demands by training a single controller only once. It relies on the Pareto front concept to achieve diverse behaviours even in non-linear task space. Finally, the optimized population of commands indicates regions of high performance in the task space.

\section{Methodology}\label{methodology}
\subsection{Problem Formulation}
We explored the problem of learning a locomotion controller for a quadrupedal robot with 12 degrees of freedom (DoF) simulated in the MuJoCo physics simulator \cite{todorov2012mujoco}. We formulated the problem as a Markov Decision Process (MDP) defined as $(S, A, p, r, \gamma)$ where $S$ is the state space, $A$ the action space, $p(s_{t+1}|a_t,s_t)$ is the transition probability function, $r : S \times A \rightarrow \mathbb{R}$ is the reward function and $\gamma$ is the discount factor. We augmented each state $s$ with the x, y linear and z angular velocities to condition the controller on velocity commands. We based the reward function on the similarity of the achieved state with the given command.

We used the Proximal Policy Optimization (PPO) \cite{schulman2017proximal} algorithm to solve this MDP. We used a n-step reward of 4 and 400 steps per simulation. Each experiment had a total of 45,000 simulations. The policy action frequency was set at 0.1 seconds, incorporating Gaussian noise with a mean of 0 and a standard deviation of 0.5. Policy updates occurred after every 100 simulations, and a reward discount ratio ($\gamma$) of 0.99 was employed. The learning rate was specified as 0.0003 and changed to 0.0001 after 30,000 simulations, with a PPO clip value of 0.2. We used fully connected networks of three layers for both actor and critic but with different hidden layer sizes, totalling 29,184 and 9,792 weights, respectively.

\subsection{Multi-Objective Fitness}
Since fitness must be positive, we divided our reward formulation into positive and negative aspects. However, for both parts, we need to normalize the reward feature. Thus, we introduced the following equation:

\begin{equation}
\label{eq:reward_norm}
r(v)= 
\begin{cases}
    (1 - norm(v))^e, & \text{if outcome should be inverted}\\
    norm(v)^e,       & \text{otherwise}
\end{cases}
\end{equation}

with $norm$ being a min-max normalization, $v$ being the feature value and $e$ an exponent to give more weight to the last increment of the feature. Then, the positive part calculates the similarity to a given command by taking the angle between the expected command coordinate and the achieved state velocities coordinate of a simulation step. Second, we take the distance between the expected command magnitude and the state velocities magnitude. Finally, we pass both values into equation \ref{eq:reward_norm} and add them together. We used the positive part in both the fitness and RL rewards. The negative part of the reward is composed of penalties for the robot's upright orientation, height and joint velocities, which we obtained by negating the result of equation \ref{eq:reward_norm}, which we then added together. These penalties encourage the robot to stay upright, avoid jumping/crawling, and move joints at nominal speeds. Adding both the positive and negative parts gave the final RL reward.

\begin{figure}
  \centering
  \includegraphics[width=0.4\linewidth]{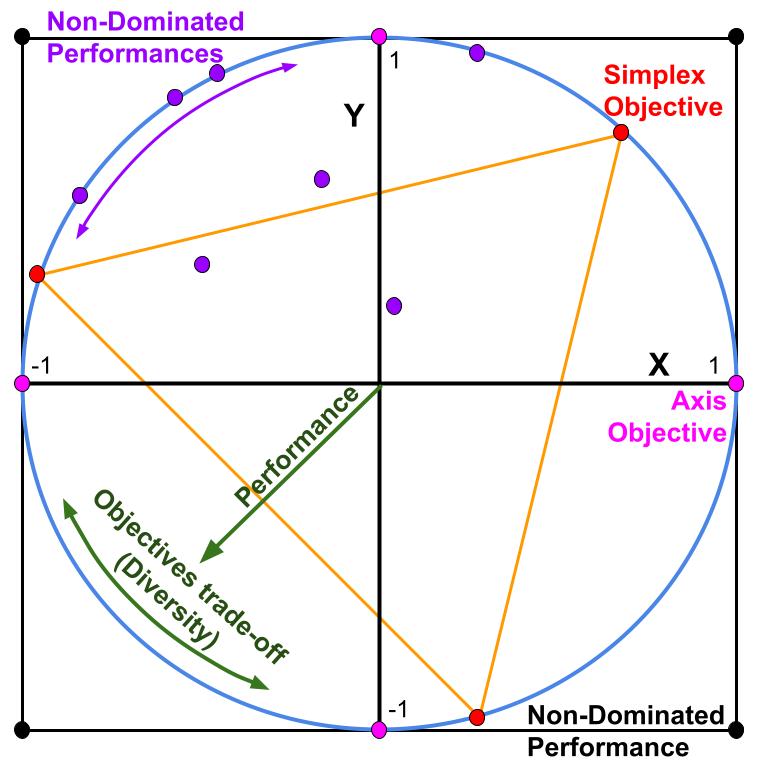}
\caption{This is a representation of the objective space in two dimensions, illustrating locomotion performance in the x and y directions. Selecting non-dominated points when there is perfect performance in at least one direction results in four points (black square) but an infinity of points when they are L2 normalized (blue circle). We use simplex (yellow triangle) vertices as objectives to keep the latter and avoid the curse of dimensionality.}
\label{fig:objective_space}
\end{figure}

\begin{minipage}{0.9\linewidth} 
\begin{algorithm}[H]
\caption{MO fitness calculation using a conversion to objectives space.}
\label{algo:fitness}
\begin{algorithmic}

\STATE \textbf{Input:} $c_e, c_p$ (command and associated performance)
\STATE \textbf{Result:} $f$ (MO fitness)
\STATE $p =$ coordinate($c_p$)
\STATE $O =$ simplex\_points()
\STATE $max_a =$ angle($o_i, o_{i+1}$)
\STATE $r =$ positive\_reward($c_e, c_p$)
\STATE $f = \emptyset$
\FOR{$o \in O$}
    \STATE $a = $ angle($p, o$)
    \STATE $a = 1 - (min(a, max_a) / max_a)$
    \STATE $f = f \cup \{a \times r\}$
\ENDFOR
\STATE return $f$
 
\end{algorithmic}
\end{algorithm}
\end{minipage}
\\
\\
To build a MO fitness, we define an objective space where commands are positioned within it by normalizing their values and magnitude within a $[-1, 1]$ range based on their maximum and minimum values. We then performed L2 normalization to project this new coordinate onto a hyper-sphere. Within this hyper-sphere, we calculate the angle between the normalized command coordinate and the vertices of an inscribed simplex, each vertex representing a distinct learning objective as seen in Figure \ref{fig:objective_space}. The angles are normalized between $[0, 1]$, subtracted from one, and multiplied by the positive part of the reward. We formally defined the conversion to a MO fitness in algorithm \ref{algo:fitness}.

As seen in Figure \ref{fig:objective_space}, the objective space places each command on the periphery of a circle (sphere or hyper-sphere based on the number of dimensions), which makes them all unique trade-offs that can dominate each other based on the learner performance calculated by the positive reward. This transformation leads to MO fitness that enables us to use any MO algorithms as ACL algorithms, specifically evolutionary ones, in this work.

\subsection{Multi-Objective Optimization}
The key idea is to use the population-based search of MO algorithms to enhance the balance between learning new commands and mastering known commands. More formally, the algorithm selects n new commands for each generation, with n being the population size. Then, an RL agent is trained m times on these commands, with m being the number of rollouts per generation. Next, for each rollout, every command is simulated and associated with its reward to update the RL agent. The latter is updated each x rollout, with x chosen based on ablation experiments. Finally, fitness for each command is calculated on the last rollout of a generation to update the MO algorithm. We repeat this process for k generations.

After each generation, the algorithm uses the population of commands to produce offspring commands during the MO update. It generates the offspring using the probabilistic uniform crossover and a mutation based on binomial probability with Gaussian noise. We used a hyper-parameter schedule to go from mainly generating random offspring to offspring close to their parent's command. The schedule is composed of three values set at generation 0, 50 and 100. The crossover probability was set to 0.5, 0.3 and 0.0, respectively. Then, the mutation probability used values of 0.8, 0.6, and 0.4. Finally, we used a mutation strength of 0.1, 0.05 and 0.01 for generations 0, 50, 100. 

\section{Results}\label{results}
We employed three metrics to evaluate our approach against the baselines. Mean Reward gauges the average effectiveness of the training strategies, highlighting performance consistency. Mean Distance assesses how precisely the robot controller matches a diverse set of commands, measuring system accuracy. Quadrant Mean Density investigate the distribution and diversity of the command populations within the algorithm, examining commands' spatial distribution in the command space. These metrics are needed to illustrate the generation of diverse, high-performing locomotion commands and the improved learning stability of our proposed approach.

We detailed the calculation of each metric used in the following paragraph. First, we computed the Mean Reward by averaging the rewards received during each of the 50 commands simulated during a rollout, with each simulation consisting of 400 steps, providing a comprehensive measure of performance over the training period. Mean Distance involves deterministically assessing the final robot controller against a specified set of commands—either a regular grid of 125 points or the retained commands in the algorithm's population—and calculating the average distance between each command and its corresponding performance outcome, thereby quantifying the precision of the control system. More formally, we calculate the following: 
\begin{equation}
\label{eq:mean_distance}
d(C_e, C_p) = \frac{1}{|C_e|}\sum^{|C_e|}_i|| c^i_p - c^i_e ||,
\end{equation}
with $c \in C$ being commands from a set of commands, $c_e$ being the command expected to be achieved and $c_p$ being the mean state performance. Lastly, Quadrant Mean Density associates each command with one of eight equal-sized quadrants, then divides the number of commands per quadrant by the expected number of commands in each quadrant when equally distributed. Specifically, we calculated it as:
\begin{equation}
\label{eq:mean_density}
s(C_p, Q) = \frac{1}{|Q|} \sum^{|Q|}_i \frac{min(|C_p \cap q_i| \ , \ |C_p| / |Q|)}{|C_p| / |Q|},
\end{equation}
with $q \in Q$ being a set of eight quadrants equally dividing the command space. These calculations provide a thorough quantitative analysis of the control system's performance and adaptability within the specified command space.

\subsection{Baseline Comparison}\label{baseline_comparison}
We employed a random command selection approach based on a uniform distribution as our baseline. Despite its simplicity, as noted in \cite{tzannetos2023proximal}, which aims for a uniform learning of the objectives, the random baseline remains robust. Additionally, work such as \cite{miki2022learning} employed a random selection of target velocities and showed impressive results even when transferred to a real-world quadrupedal robot. We also compare our approach to ADR from \cite{akkaya2019solving}, which progressively increases task difficulties by expanding its sampling range of the task pool. It successfully learned dexterous hand manipulations, a complex, open-ended problem, by sampling continuous task parameters, making it a relevant baseline. We compared NSGA-II and MOEA/D as two variants of our MOL approach against these baselines. While both algorithms aim to identify the optimal Pareto set, NSGA-II utilizes a non-dominated sorting algorithm, whereas MOEA/D is founded on objective decomposition to generate multiple single objectives. Note that we compared the weighted sum and Tchebycheff aggregation functions and chose the latter for our version of MOEA/D.

%%%%%%%%%% Explanation of population %%%%%%%%%%
At the end of an optimization, both algorithms yield their populations representing the optimal trade-offs among objectives. Regarding ADR and the Random baseline, while they randomly select commands, we added the selection made by NSGA-II to preserve the best-performing objectives trade-off observed, but only for comparison purposes.

%%%%%%%%%% Scenarios %%%%%%%%%%
To rigorously assess the capabilities and limitations of the algorithms, four distinct simulation scenarios were developed: nominal, limited, back and run. In the nominal scenario, the quadrupedal robot operates under optimal conditions with complete freedom in its degrees of movement. The limited scenario constrains the robot's hip movements to a single degree, preventing lateral mobility. The back scenario restricts each actuator in the robot's hind legs to a single degree of movement, essentially removing the functionality of these legs. Lastly, the run scenario only enables individual action from the policy of 0.6 magnitudes or more. Otherwise, it is set to zero, forcing the policy toward running behaviours. We designed these scenarios to highlight the strengths and weaknesses of the algorithms across different operational constraints.

%%%%%%%%%% Accuracy - Generalization %%%%%%%%%%

\begin{figure*}
\begin{subfigure}{\textwidth}
  \centering
  \includegraphics[width=0.45\linewidth]{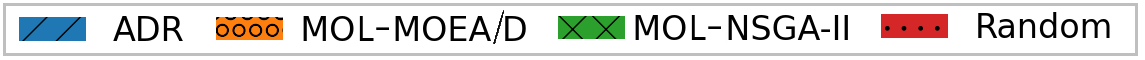}
  \label{fig:pop_dist}
\end{subfigure}
  \begin{subfigure}{.5\textwidth}
  \centering
  \includegraphics[width=0.75\linewidth]{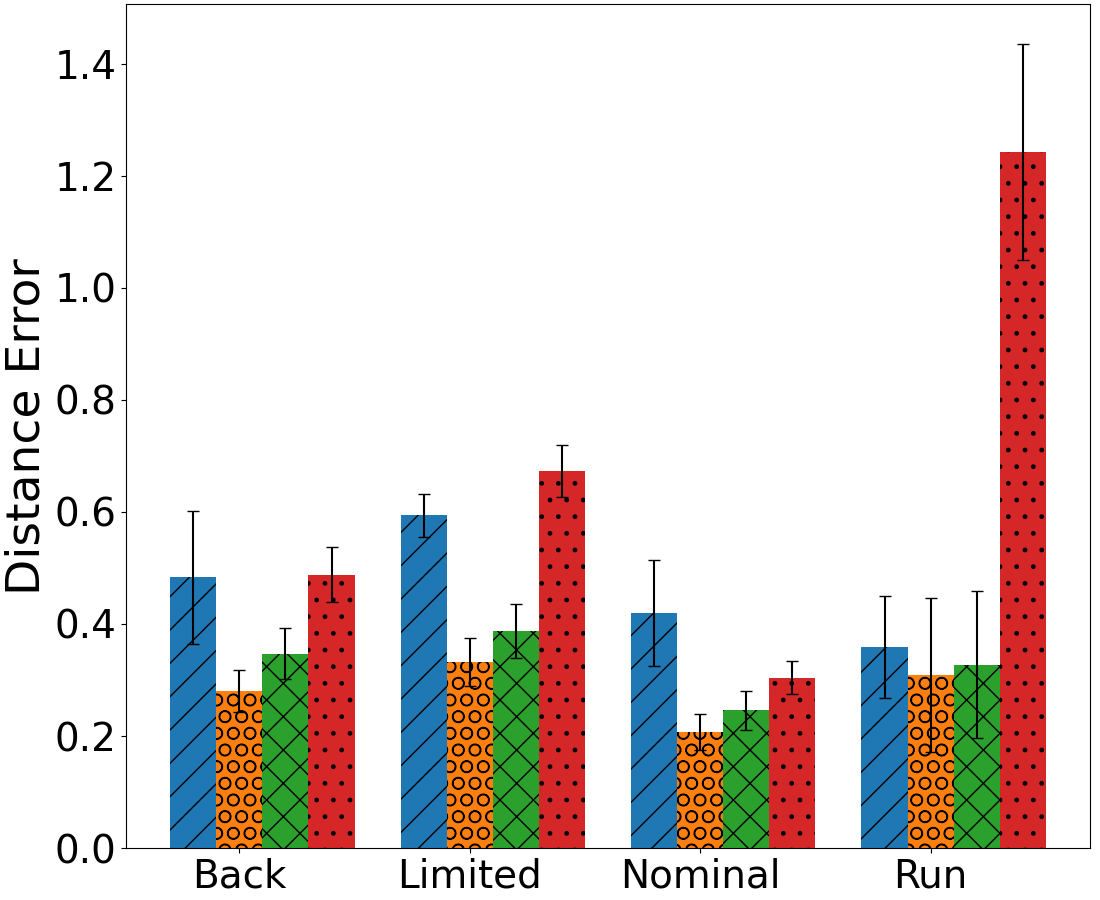}
  \caption{Population set of commands.}
  \label{fig:pop_dist}
\end{subfigure}%
\begin{subfigure}{.5\textwidth}
  \centering
  \includegraphics[width=0.75\linewidth]{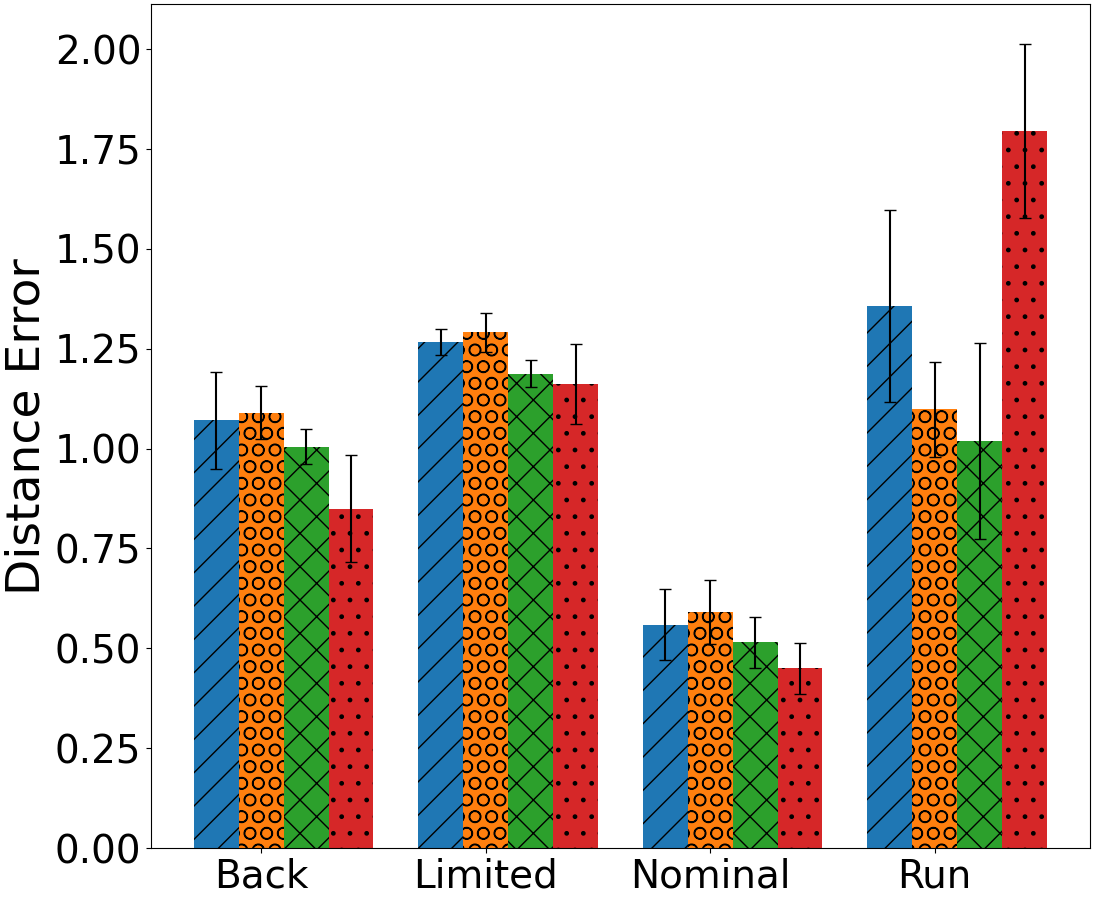}
  \caption{Test set of commands.}
  \label{fig:test_dist}
\end{subfigure}
  \caption{Comparison of the approaches on different operational constraints based on their mean performance distance to a set of desired commands. Lower is better; the cap bar represents one standard deviation from the trial's mean. Results are from ten trials for each approach in each scenario.}
\label{fig:distance_comparison}
\end{figure*}

The results in Figure \ref{fig:distance_comparison} show that MOL variants achieve comparable or better performance on the set of test commands. For example, in the run scenario, a t-test between MOEA/D and ADR gives a p-value of 0.0127 with a 0.259 mean reduction in distance error for MOEA/D. Furthermore, the performance on the set of population commands surpasses that on the uniformly distributed commands in every case. Also, MO optimization algorithms show even lower errors than baselines in every scenario. The Limited scenario strongly illustrated this advantage with a t-test between MOEA/D and ADR, giving a p-value of p \textless \ 0.001 with a 0.263 mean reduction in distance error for MOEA/D. These results illustrate the capacity of MO algorithms to adapt to difficult scenarios. Moreover, the similar performances of all approaches on the nominal scenario with the set of test commands show that our approach also does not lose performance in the default use case. These results show that our approach improves performance on a valuable set of diverse commands while retaining good generalization capabilities for all possible commands.

\begin{figure*}
\begin{subfigure}{\textwidth}
  \centering
  \includegraphics[width=0.45\linewidth]{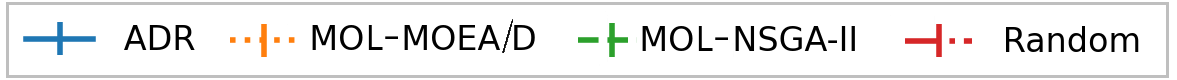}
  \label{fig:plots_legend}
\end{subfigure}
  \begin{subfigure}{.259\textwidth}
  \centering
  \includegraphics[width=0.99\linewidth]{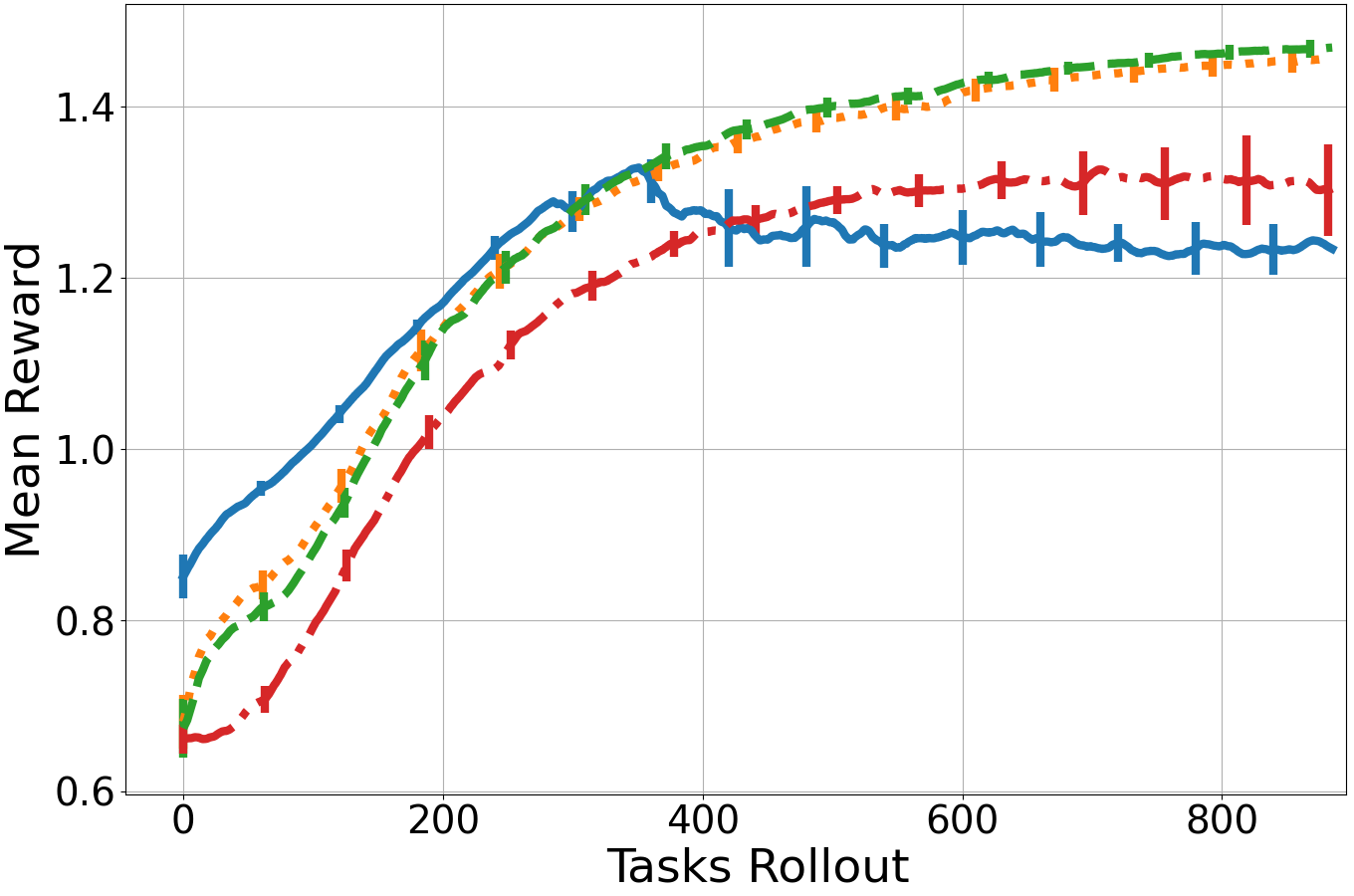}
  \caption{Nominal}
  \label{fig:nominal_reward}
\end{subfigure}%
\begin{subfigure}{.247\textwidth}
  \centering
  \includegraphics[width=0.99\linewidth]{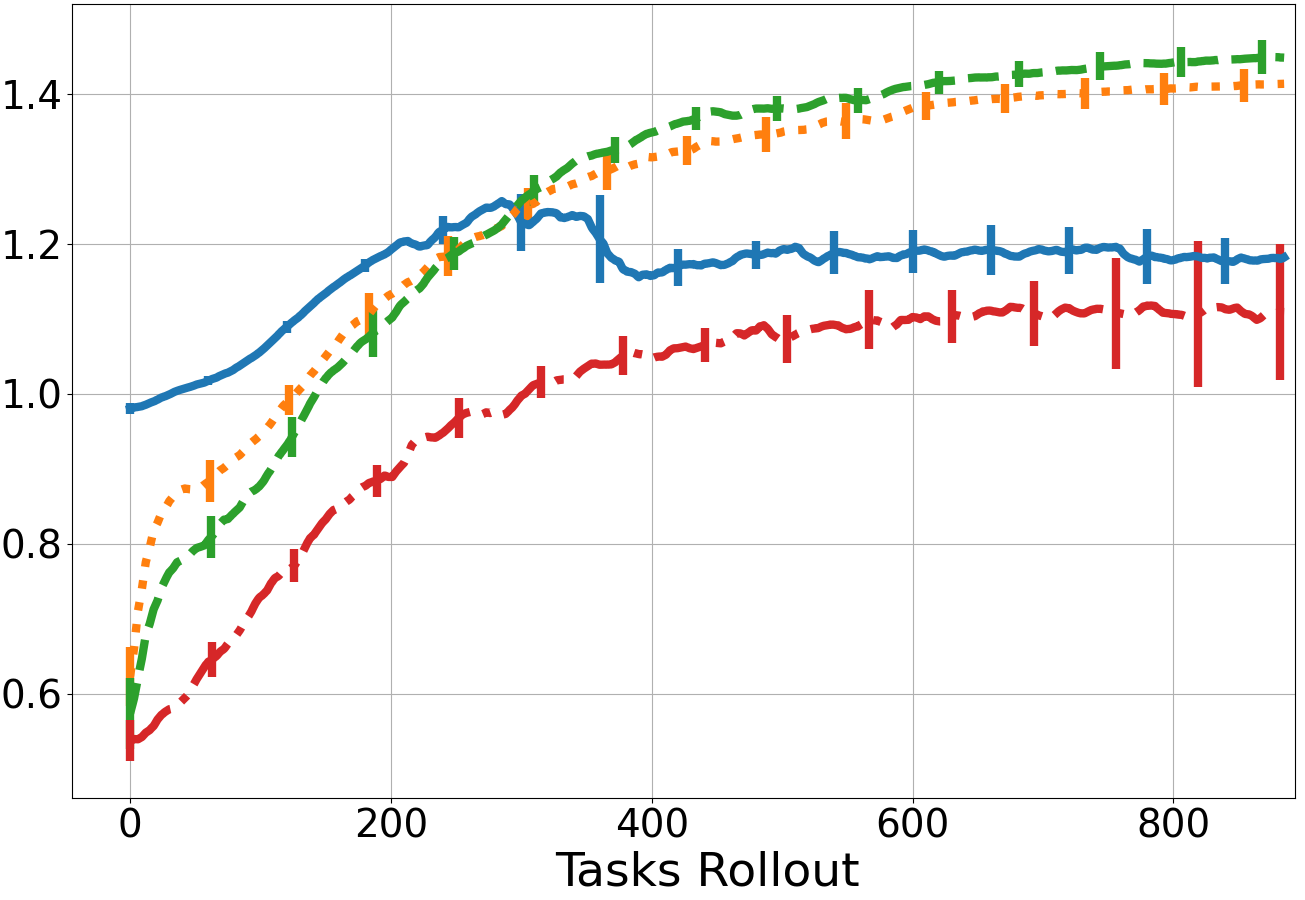}
  \caption{Limited}
  \label{fig:limited_reward}
\end{subfigure}
\begin{subfigure}{.247\textwidth}
  \centering
  \includegraphics[width=0.99\linewidth]{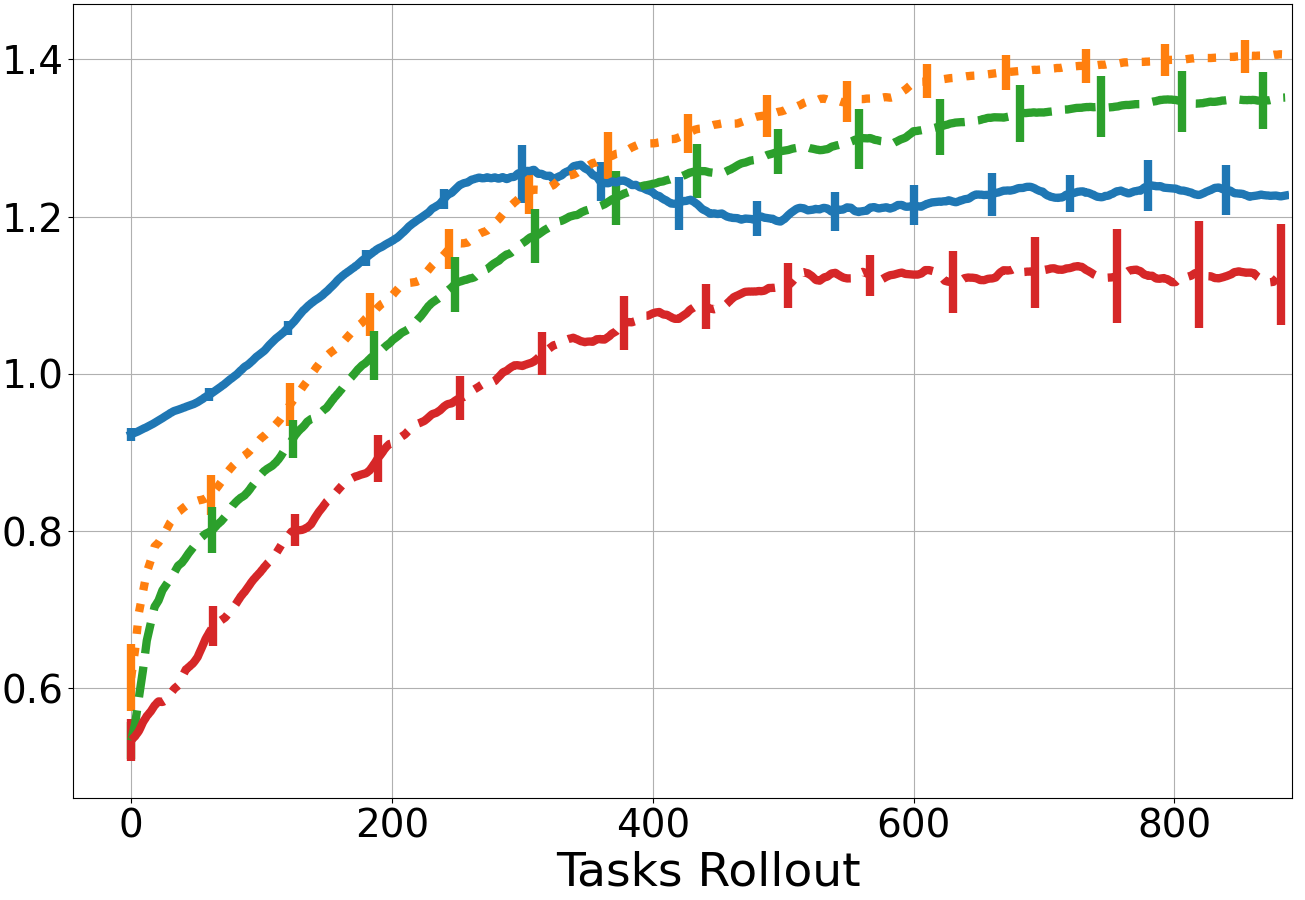}
  \caption{Back}
  \label{fig:back_reward}
\end{subfigure}%
\begin{subfigure}{.247\textwidth}
  \centering
  \includegraphics[width=0.99\linewidth]{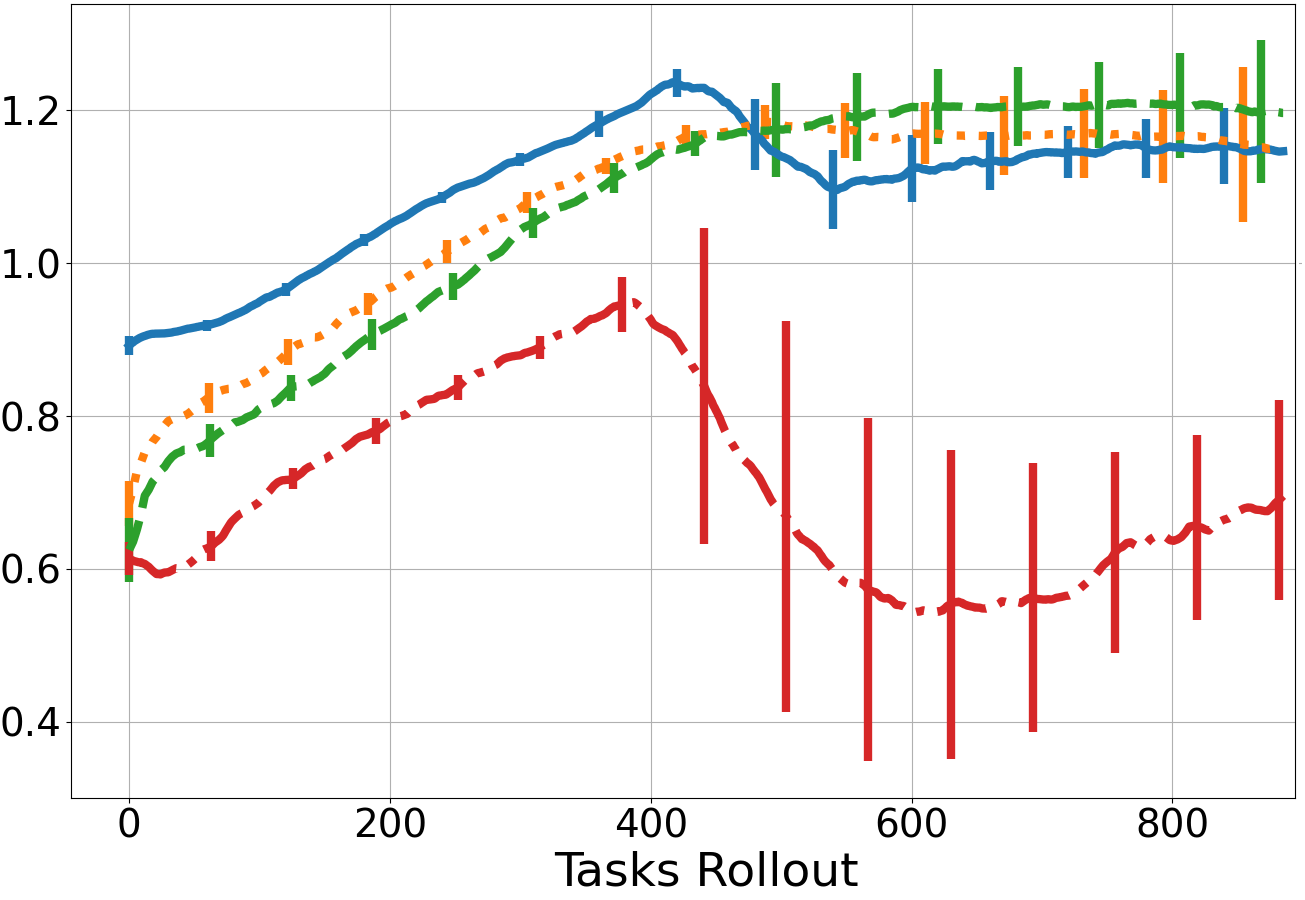}
  \caption{Run}
  \label{fig:run_reward}
\end{subfigure}
  \caption{Comparison of training mean reward for all operational constraints. The mean reward was calculated from every 400 steps of the 50 sampled commands. Error bar represents one standard deviation from the trials mean. Constant progression and lower error bar indicate better training stability. Results are from ten trials for each approach in each scenario.}
\label{fig:reward_comparison}
\end{figure*}

%%%%%%%%%% Stability %%%%%%%%%%
The results in Figure \ref{fig:reward_comparison} illustrate that using a curriculum improves training stability. The latter is visible with NSGA-II, MOEA/D and ADR displaying smaller standard deviations than the Random baseline, indicating better reproducibility. Moreover, our MOL approach shows even more stability with the continuous monotonic increase of its mean reward, which indicates continuous improvement, except for the run scenario, where the reward became saturated at the end. The large error bars of the Random algorithm in figure \ref{fig:run_reward} are explained by the controller either learning to follow commands or failing to learn and mostly remaining in place.

%%%%%%%%%% Diversity %%%%%%%%%%
It could be conjectured that the improved training stability of MOL is due to the algorithms' use of a clustered population. However, all approaches show similar performance when looking at the Mean Quadrant Density detailed in Figure \ref{fig:mean_density}. This performance indicates that our approach has improved stability while maintaining a population as diverse as its competitors and can achieve similar or improved performance. In other words, despite the scenarios constraints, MOL variants can focus on both the performing regions of the task space and on the regions at the edge of the constraints, exploring most of the achievable diversity in the task space.

%%%%% NSGA-II mutation %%%%%
\subsection{Mutation Ablation}\label{mutation_ablation}
\begin{figure*}

\centering
\begin{subfigure}{0.5\textwidth}
  \centering
  \includegraphics[width=0.9\linewidth]{figures/Bars_Legend.png}
\end{subfigure}%
\begin{subfigure}{0.5\textwidth}
  \centering
  \includegraphics[width=0.55\linewidth]{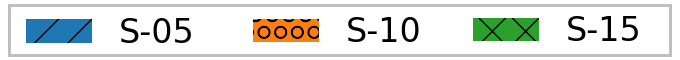}
\end{subfigure}
\begin{subfigure}{0.5\textwidth}
  \centering
  \includegraphics[width=0.75\linewidth]{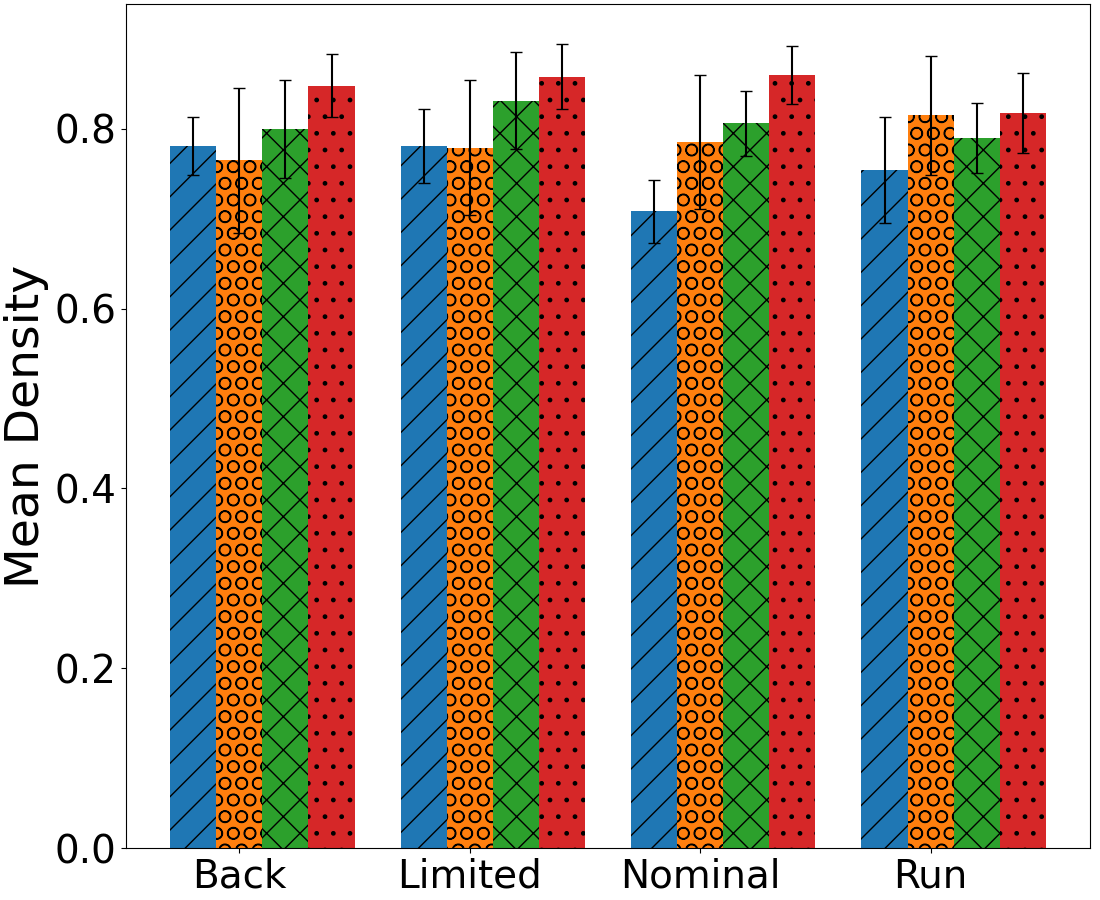}
  \caption{Mean Quadrant Density}
  \label{fig:mean_density}
\end{subfigure}%
\begin{subfigure}{0.5\textwidth}
  \centering
  \includegraphics[width=0.7\linewidth]{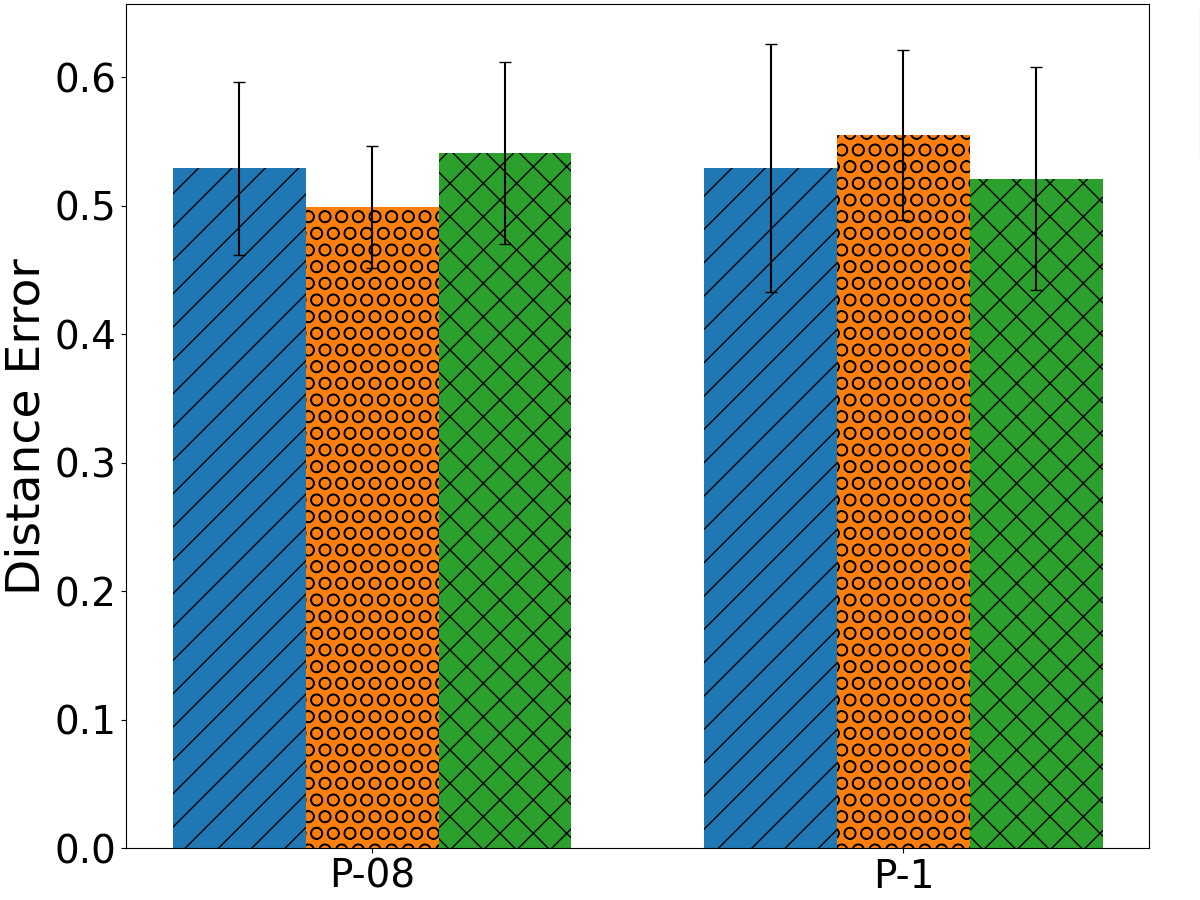}
  \caption{Mutation ablation}
  \label{fig:mutation_ablation}
\end{subfigure}

\caption{a) Mean density for all eight quadrants of the task space per approach per operational constraints. Higher is better, and the cap bar represents one standard deviation from the trial's mean. Results are from ten trials for each approach in each scenario. b) Ablation study of different schedules for the mutation probability (P-08, P-1) and strength (S-05, S-10, S-15) using NSGA-II. Results are from the nominal scenario and based on the set of test commands. Lower is better, and the cap bar is one standard deviation from the mean of five trials.}
\end{figure*}

We investigated the mutation probability and strength within the NSGA-II algorithm to select the most performing values. These parameters are critical for the exploration of the task space by MO algorithms. We tested two mutation probabilities and three mutation strengths schedules (value set at generation 0, 50 and 100). Figure \ref{fig:mutation_ablation} shows P-08 ([0.8, 0.6, 0.4]), P-1 ([1.0, 0.8, 0.6]) as the probabilities schedules and S-05 ([0.05, 0.025, 0.005]), S-10 ([0.1, 0.05, 0.01]), S-15 ([0.15, 0.075, 0.03]) as the strength schedules. Note that strength values are relative to a velocity of 1.0.

The results display the Mean Distance on the set of test commands and indicate that P-08 paired with S-10 yields the most favourable outcomes. This particular parameter combination also proved competitive on the population set. For computation reasons, we used the same parameters schedule for MOEA/D. Due to the addition of hyperparameters for MOEA/D when using the standard version with differential evolution, we used the probabilistic uniform crossover operation for MOEA/D with the same hyperparameters as NSGA-II chosen in preliminary experiments.

\subsection{Rollout And Update Ablation}\label{rollout-update_ablation}

%%%%% NSGA-II rollout-update %%%%%
\begin{figure*}
\begin{subfigure}{\textwidth}
  \centering
  \includegraphics[width=0.15\linewidth]{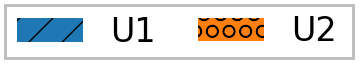}
  \label{fig:roll_up_legend}
\end{subfigure}
\begin{subfigure}{.33\textwidth}
  \centering
  \includegraphics[width=1.0\linewidth]{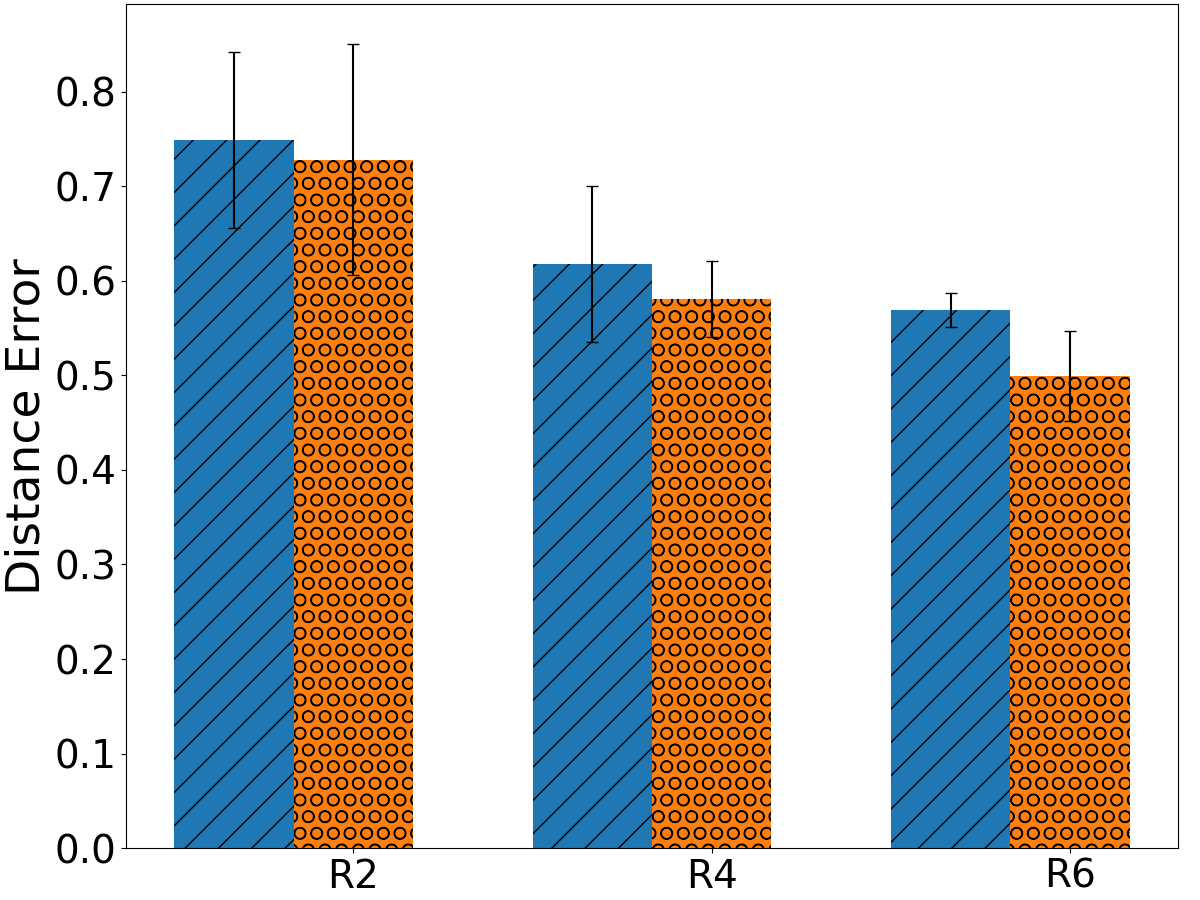}
  \caption{NSGA-II}
  \label{fig:nsga2_roll-up_test}
\end{subfigure}
\begin{subfigure}{.33\textwidth}
  \centering
  \includegraphics[width=1.0\linewidth]{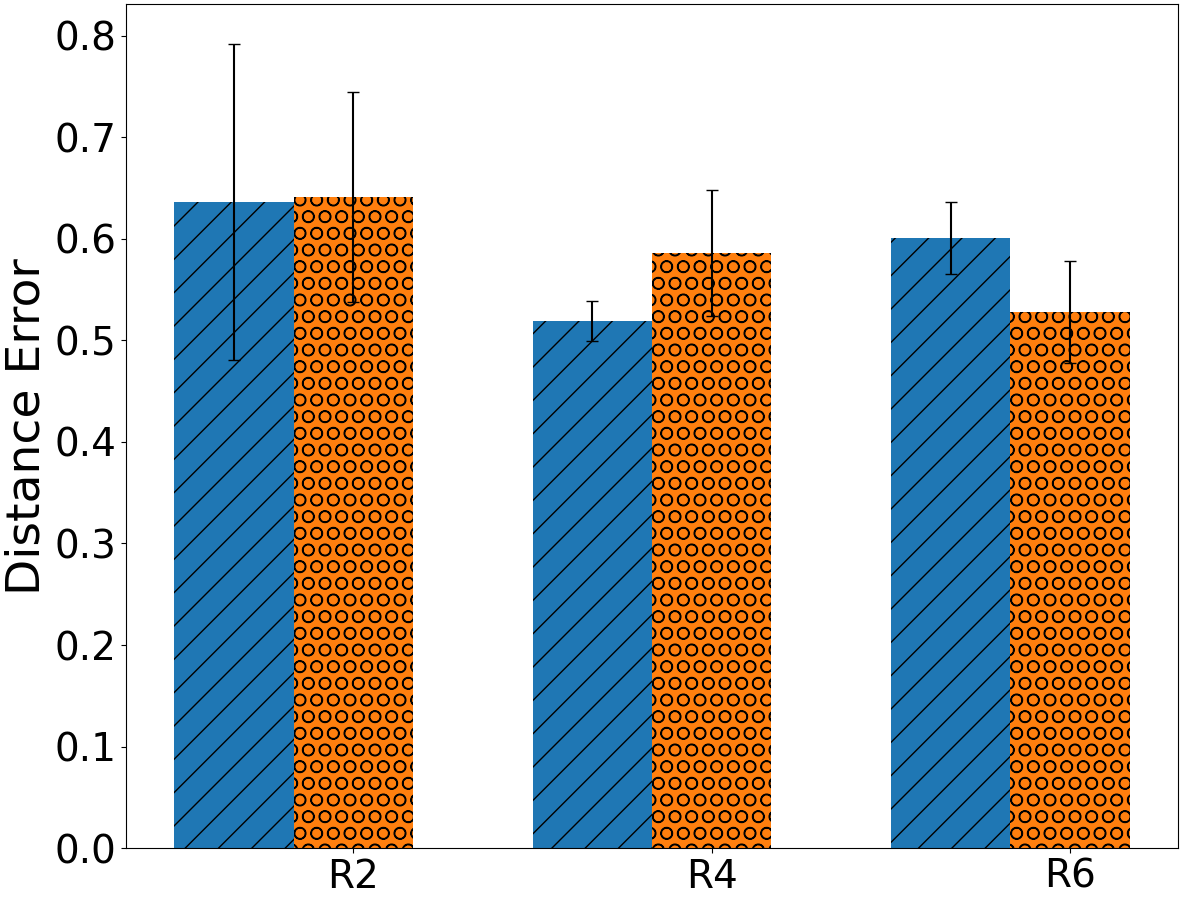}
  \caption{ADR}
  \label{fig:adr_roll-up_pop}
\end{subfigure}
\begin{subfigure}{.33\textwidth}
  \centering
  \includegraphics[width=1.0\linewidth]{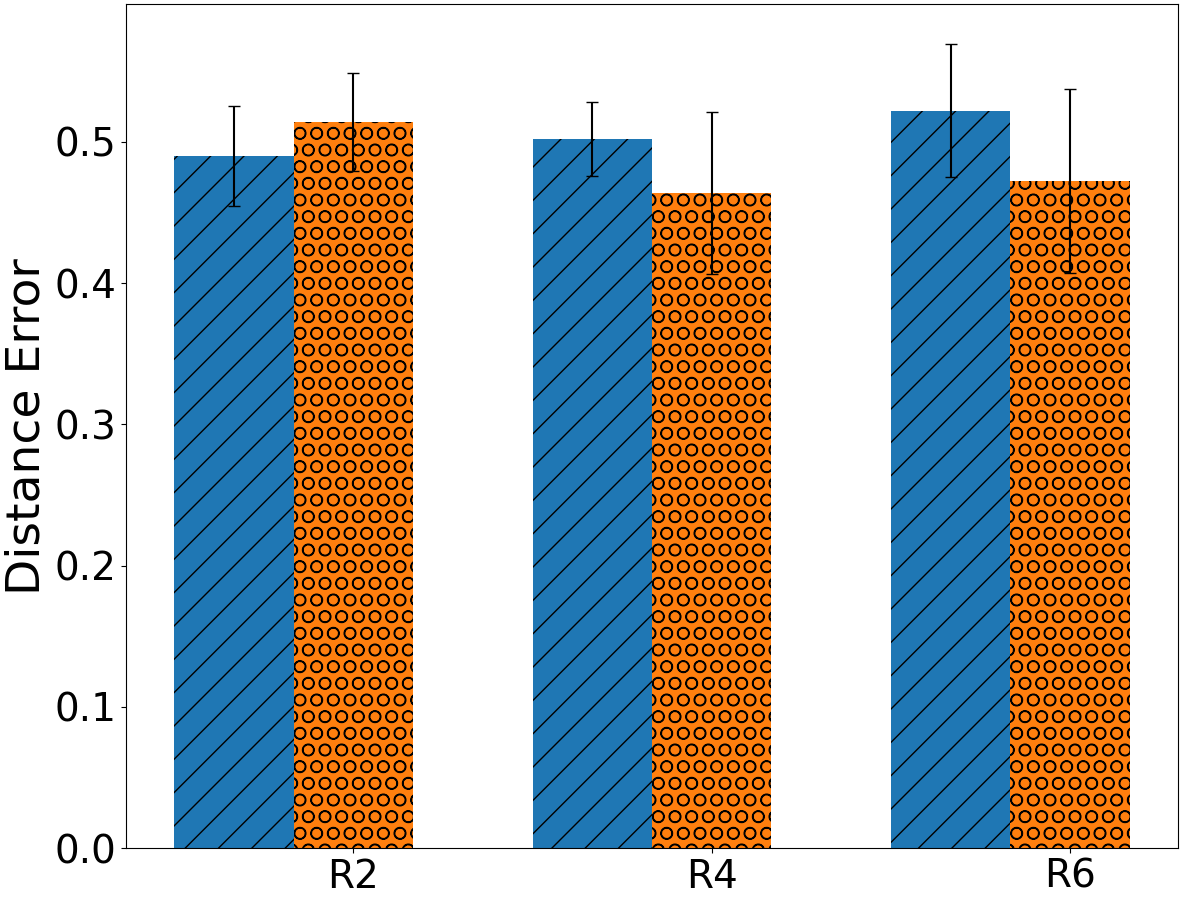}
  \caption{Random}
  \label{fig:random_roll-up_pop}
\end{subfigure}
  \caption{Ablation study of the number of rollouts (R2, R4, R6) and RL update per rollout (U1, U2) per generation. Results are from the nominal scenario and based on the set of test commands. Lower is better, and the cap bar is one standard deviation from the mean of five trials.}
\label{fig:roll-up_ablation}
\end{figure*}

We investigated the interaction between the number of rollouts 2, 4 and 6 (R2, R4, R6) per generation and a frequency of 1 or 2 (U1, U2) RL updates per rollout to optimize the generation length and update periodicity. These parameters are essential to balance the number of different commands and the number of steps per command seen at each RL update. Figure \ref{fig:roll-up_ablation} shows the performance on the set of test commands since it indicates the algorithm's ability to generalize. The results for NSGA-II suggest that the R6 and U2 combination—yield the most advantageous outcomes. We used this combination of parameters in both NSGA-II and MOEA/D for computation reasons. Since the combination of R6 and U2 is also competitive on the set of test commands for ADR and Random and performs best for both on the set of population commands, we chose this combination for the baselines.

\section{Conclusion}\label{conclusion}
To conclude, we proposed the novel MOL approach for ACL. We showed that our approach has improved stability while maintaining a population as diverse as its competitors and can achieve similar or improved performance for quadrupedal locomotion. As such, our approach achieved 19\% and 44\% fewer errors against our best baseline algorithm in difficult scenarios based on a uniform and tailored evaluation respectively.

Despite these performances, MOL still relies on experimentally chosen schedules of hyperparameters, and while scaling linearly in the number of objectives, MO algorithms are known to handle up to four objectives optimally. However, it opens up future research directions, such as dynamically adjusting mutation and crossover parameters, replacing the MO algorithms with many-objective algorithms, scaling up the population size and applying the approach to new open-ended problems, such as procedurally generated terrains based on real parameters. Transfer from simulation to real-world robots could also be attempted using a teacher-student or domain randomization approach.

\bibliography{Bibliography}

\begin{thebibliography}{10}

\bibitem{akkaya2019solving}
Ilge Akkaya, Marcin Andrychowicz, Maciek Chociej, Mateusz Litwin, Bob McGrew, Arthur Petron, Alex Paino, Matthias Plappert, Glenn Powell, Raphael Ribas, et~al.
\newblock Solving rubik's cube with a robot hand.
\newblock {\em arXiv preprint arXiv:1910.07113}, 2019.

\bibitem{andrychowicz2021matters}
Marcin Andrychowicz, Anton Raichuk, Piotr Sta{\'n}czyk, Manu Orsini, Sertan Girgin, Rapha{\"e}l Marinier, L{\'e}onard Hussenot, Matthieu Geist, Olivier Pietquin, Marcin Michalski, et~al.
\newblock What matters in on-policy reinforcement learning? a large-scale empirical study.
\newblock In {\em ICLR 2021-Ninth International Conference on Learning Representations}, 2021.

\bibitem{bengio2009curriculum}
Yoshua Bengio, J{\'e}r{\^o}me Louradour, Ronan Collobert, and Jason Weston.
\newblock Curriculum learning.
\newblock In {\em Proceedings of the 26th annual international conference on machine learning}, pages 41--48, 2009.

\bibitem{chignoli2021humanoid}
Matthew Chignoli, Donghyun Kim, Elijah Stanger-Jones, and Sangbae Kim.
\newblock The mit humanoid robot: Design, motion planning, and control for acrobatic behaviors.
\newblock In {\em 2020 IEEE-RAS 20th International Conference on Humanoid Robots (Humanoids)}. IEEE, 2021.

\bibitem{pmlr-v80-florensa18a}
Carlos Florensa, David Held, Xinyang Geng, and Pieter Abbeel.
\newblock Automatic goal generation for reinforcement learning agents.
\newblock In {\em Proceedings of the 35th International Conference on Machine Learning}, volume~80 of {\em Proceedings of Machine Learning Research}, pages 1515--1528. PMLR, 10--15 Jul 2018.

\bibitem{haarnoja2023learning}
Tuomas Haarnoja, Ben Moran, Guy Lever, Sandy~H Huang, Dhruva Tirumala, Jan Humplik, Markus Wulfmeier, Saran Tunyasuvunakool, Noah~Y Siegel, Roland Hafner, et~al.
\newblock Learning agile soccer skills for a bipedal robot with deep reinforcement learning.
\newblock {\em Science Robotics}, 2024.

\bibitem{huizinga2022evolving}
Joost Huizinga and Jeff Clune.
\newblock Evolving multimodal robot behavior via many stepping stones with the combinatorial multiobjective evolutionary algorithm.
\newblock {\em Evolutionary Computation}, 30(2):131--164, 2022.

\bibitem{mehta2020active}
Bhairav Mehta, Manfred Diaz, Florian Golemo, Christopher~J Pal, and Liam Paull.
\newblock Active domain randomization.
\newblock In {\em Conference on Robot Learning}, pages 1162--1176. PMLR, 2020.

\bibitem{miki2022learning}
Takahiro Miki, Joonho Lee, Jemin Hwangbo, Lorenz Wellhausen, Vladlen Koltun, and Marco Hutter.
\newblock Learning robust perceptive locomotion for quadrupedal robots in the wild.
\newblock {\em Science Robotics}, 7(62):eabk2822, 2022.

\bibitem{ijcai2020p671}
Rémy Portelas, Cédric Colas, Lilian Weng, Katja Hofmann, and Pierre-Yves Oudeyer.
\newblock Automatic curriculum learning for deep rl: A short survey.
\newblock In {\em Proceedings of the Twenty-Ninth International Joint Conference on Artificial Intelligence, {IJCAI-20}}, pages 4819--4825. IJCAI Organization, 7 2020.

\bibitem{DBLP-conf-iclr-RacaniereLSRFL20}
S{\'{e}}bastien Racani{\`{e}}re, Andrew~K. Lampinen, Adam Santoro, David~P. Reichert, Vlad Firoiu, and Timothy~P. Lillicrap.
\newblock Automated curriculum generation through setter-solver interactions.
\newblock In {\em 8th International Conference on Learning Representations, {ICLR}}. OpenReview.net, 2020.

\bibitem{schulman2017proximal}
John Schulman, Filip Wolski, Prafulla Dhariwal, Alec Radford, and Oleg Klimov.
\newblock Proximal policy optimization algorithms.
\newblock {\em arXiv preprint arXiv:1707.06347}, 2017.

\bibitem{seita2019zpd}
Daniel Seita, David Chan, Roshan Rao, Chen Tang, Mandi Zhao, and John Canny.
\newblock Zpd teaching strategies for deep reinforcement learning from demonstrations.
\newblock {\em arXiv preprint arXiv:1910.12154}, 2019.

\bibitem{shabani2010vygotsky}
Karim Shabani, Mohamad Khatib, and Saman Ebadi.
\newblock Vygotsky's zone of proximal development: Instructional implications and teachers' professional development.
\newblock {\em English language teaching}, 3(4):237--248, 2010.

\bibitem{sukhbaatar2018intrinsic}
Sainbayar Sukhbaatar, Zeming Lin, Ilya Kostrikov, Gabriel Synnaeve, Arthur Szlam, and Rob Fergus.
\newblock Intrinsic motivation and automatic curricula via asymmetric self-play.
\newblock In {\em International Conference on Learning Representations}, 2018.

\bibitem{todorov2012mujoco}
Emanuel Todorov, Tom Erez, and Yuval Tassa.
\newblock Mujoco: A physics engine for model-based control.
\newblock In {\em 2012 IEEE/RSJ International Conference on Intelligent Robots and Systems}, pages 5026--5033. IEEE, 2012.

\bibitem{tzannetos2023proximal}
George Tzannetos, B{\'a}rbara Gomes~Ribeiro, Parameswaran Kamalaruban, and Adish Singla.
\newblock Proximal curriculum for reinforcement learning agents.
\newblock {\em Transactions on Machine Learning Research}, 2023(5):1--21, 2023.

\bibitem{wang2019paired}
Rui Wang, Joel Lehman, Jeff Clune, and Kenneth~O Stanley.
\newblock Paired open-ended trailblazer (poet): Endlessly generating increasingly complex and diverse learning environments and their solutions.
\newblock {\em arXiv preprint arXiv:1901.01753}, 2019.

\bibitem{wang2023toward}
William~Wei Wang, Dongqi Han, Xufang Luo, Yifei Shen, Charles Ling, Boyu Wang, and Dongsheng Li.
\newblock Toward open-ended embodied tasks solving.
\newblock {\em arXiv preprint arXiv:2312.05822}, 2023.

\end{thebibliography}

\end{document}